# Parsing with the Shortest Derivation

**Rens Bod**

Informatics Research Institute, University of Leeds, Leeds LS2 9JT, &
Institute for Logic, Language and Computation, University of Amsterdam
rens@scs.leeds.ac.uk

## Abstract

Common wisdom has it that the bias of stochastic grammars in favor of shorter derivations of a sentence is harmful and should be redressed. We show that the common wisdom is wrong for stochastic grammars that use elementary trees instead of context-free rules, such as Stochastic Tree-Substitution Grammars used by Data-Oriented Parsing models. For such grammars a *non*-probabilistic metric based on the shortest derivation outperforms a probabilistic metric on the ATIS and OVIS corpora, while it obtains competitive results on the Wall Street Journal (WSJ) corpus. This paper also contains the first published experiments with DOP on the WSJ.

## 1. Introduction

A well-known property of stochastic grammars is their propensity to assign higher probabilities to shorter derivations of a sentence (cf. Chitrao & Grishman 1990; Magerman & Marcus 1991; Briscoe & Carroll 1993; Charniak 1996). This propensity is due to the probability of a derivation being computed as the product of the rule probabilities, and thus shorter derivations involving fewer rules tend to have higher probabilities, almost regardless of the training data. While this bias may seem interesting in the light of the principle of cognitive economy, shorter derivations generate smaller parse trees (consisting of fewer nodes) which are not warranted by the correct parses of sentences. Most systems therefore redress this bias, for instance by normalizing the derivation probability (see Caraballo & Charniak 1998).

However, for stochastic grammars that use elementary trees instead of context-free rules, the propensity to assign higher probabilities to shorter derivations does not necessarily lead to a bias in favor of smaller parse trees, because elementary trees may differ in size and lexicalization. For Stochastic Tree-Substitution Grammars (STSG) used by Data-Oriented Parsing (DOP) models, it has been observed that the shortest derivation of a sentence consists of the *largest* subtrees seen in a treebank that generate that sentence (cf. Bod 1992, 98). We may therefore wonder whether for STSG the bias in favor of shorter derivations is perhaps beneficial rather than harmful.

To investigate this question we created a new STSG-DOP model which uses this bias as a feature. This *non*-probabilistic DOP model parses each sentence by returning its shortest derivation (consisting of the fewest subtrees seen in the corpus). Only if there is more than one shortest derivation the model backs off to a frequency ordering of the corpus-subtrees and chooses the shortest derivation with most highest ranked subtrees. We compared this non-probabilistic DOP model against the probabilistic DOP model (which estimates the most probable parse for each sentence) on three different domains: the Penn ATIS treebank (Marcus et al. 1993), the Dutch OVIS treebank (Bonnema et al. 1997) and the Penn Wall Street Journal (WSJ) treebank (Marcus et al. 1993). Surprisingly, the non-probabilistic DOP model outperforms the probabilistic DOP model on both the ATIS and OVIS treebanks, while it obtains competitive results on the WSJ treebank. We conjecture that any stochastic grammar which uses units of flexible size can be turned into an accurate non-probabilistic version.

The rest of this paper is organized as follows: we first explain both the probabilistic and non-probabilistic DOP model. Next, we go into the computational aspects of these models, and finally we compare the performance of the models on the three treebanks.

## 2. Probabilistic vs. Non-Probabilistic Data-Oriented Parsing

Both probabilistic and non-probabilistic DOP are based on the DOP model in Bod (1992) which extracts a Stochastic Tree-Substitution Grammar from a treebank ("STSG-DOP").[1] STSG-DOP uses subtrees

---

[1] Note that the DOP-approach of extracting grammars from corpora has been applied to a wide variety of other grammatical frameworks, including Tree-Insertion Grammar

from parse trees in a corpus as elementary trees, and leftmost-substitution to combine subtrees into new trees. As an example, consider a very simple corpus consisting of only two trees (we leave out some subcategorizations to keep the example simple):

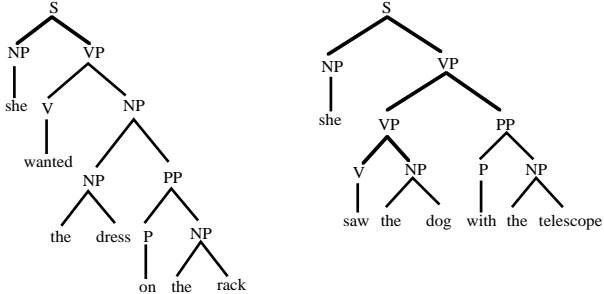

Figure 1. A simple corpus of two trees.

A new sentence such as *She saw the dress with the telescope* can be parsed by combining subtrees from this corpus by means of leftmost-substitution (indicated as ○):

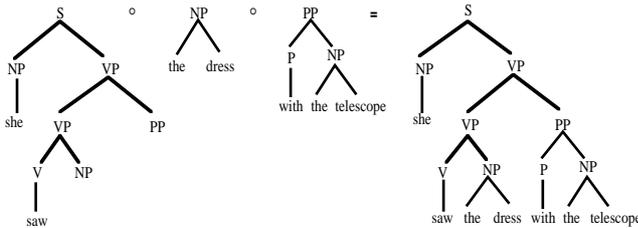

Figure 2. Derivation and parse tree for the sentence *She saw the dress with the telescope*

Note that other derivations, involving different subtrees, may yield the same parse tree; for instance:

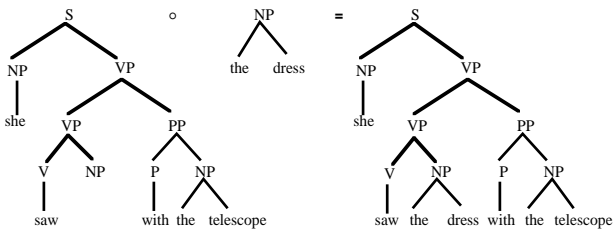

Figure 3. Different derivation yielding the same parse tree for *She saw the dress with the telescope*

Note also that, given this example corpus, the sentence we considered is ambiguous; by combining

(Hoogweg 2000), Tree-Adjoining Grammar (Neumann 1998), Lexical-Functional Grammar (Bod & Kaplan 1998; Way 1999; Bod 2000a), Head-driven Phrase Structure Grammar (Neumann & Flickinger 1999), and Montague Grammar (van den Berg et al. 1994; Bod 1998). For the relation between DOP and Memory-Based Learning, see Daelemans (1999).

other subtrees, a different parse may be derived, which is analogous to the first rather than the second corpus sentence:

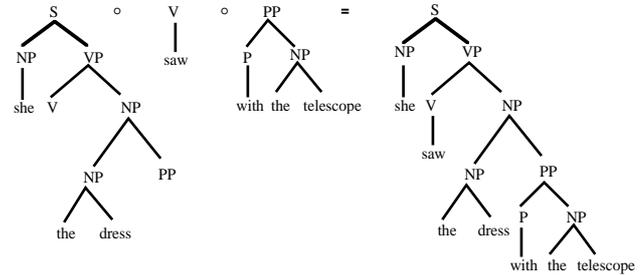

Figure 4. Different derivation yielding a different parse tree for *She saw the dress with the telescope*

The probabilistic and non-probabilistic DOP models differ in the way they define the best parse tree of a sentence. We now discuss these models separately.

### 2.1 The probabilistic DOP model

The probabilistic DOP model introduced in Bod (1992, 93) computes the most probable parse tree of a sentence from the normalized subtree frequencies in the corpus. The probability of a subtree $t$ is estimated as the number of occurrences of $t$ seen in the corpus, divided by the total number of occurrences of corpus-subtrees that have the same root label as $t$. Let $|t|$ return the number of occurrences of $t$ in the corpus and let $r(t)$ return the root label of $t$ then: $P(t) = |t| / \sum_{t': r(t')=r(t)} |t'|$.[2] The probability of a derivation is computed as the product of the probabilities of the subtrees involved in it. The probability of a parse tree is computed as the sum of the probabilities of all distinct derivations that produce that tree. The parse tree with the highest

---

[2] It should be stressed that there may be several other ways to estimate subtree probabilities in DOP. For example, Bonnema et al. (1999) estimate the probability of a subtree as the probability that it has been involved in the derivation of a corpus tree. It is not yet known whether this alternative probability model outperforms the model in Bod (1993). Johnson (1998) pointed out that the subtree estimator in Bod (1993) yields a statistically inconsistent model. This means that as the training corpus increases the corresponding sequences of probability distributions do not converge to the true distribution that generated the training data. Experiments with a *consistent* maximum likelihood estimator (based on the inside-outside algorithm in Lari and Young 1990), leads however to a significant decrease in parse accuracy on the ATIS and OVIS corpora. This indicates that statistical consistency does not necessarily lead to better performance.

probability is defined as the best parse tree of a sentence.

The probabilistic DOP model thus considers counts of subtrees of a wide range of sizes in computing the probability of a tree: everything from counts of single-level rules to counts of entire trees.

### 2.2 The non-probabilistic DOP model

The non-probabilistic DOP model uses a rather different definition of the best parse tree. Instead of computing the most probable parse of a sentence, it computes the parse tree which can be generated by the fewest corpus-subtrees, i.e., by the shortest derivation independent of the subtree probabilities. Since subtrees are allowed to be of arbitrary size, the shortest derivation typically corresponds to the parse tree which consists of *largest* possible corpus-subtrees, thus maximizing syntactic context. For example, given the corpus in Figure 1, the best parse tree for *She saw the dress with the telescope* is given in Figure 3, since that parse tree can be generated by a derivation of only two corpus-subtrees, while the parse tree in Figure 4 needs at least three corpus-subtrees to be generated. (Interestingly, the parse tree with the shortest derivation in Figure 3 is also the most probable parse tree according to probabilistic DOP for this corpus, but this need not always be so. As mentioned, the probabilistic DOP model has already a bias to assign higher probabilities to parse trees that can be generated by shorter derivations. The non-probabilistic DOP model makes this bias absolute.)

The shortest derivation may not be unique: it may happen that different parses of a sentence are generated by the same minimal number of corpus-subtrees. In that case the model backs off to a frequency ordering of the subtrees. That is, all subtrees of each root label are assigned a rank according to their frequency in the corpus: the most frequent subtree (or subtrees) of each root label get rank 1, the second most frequent subtree gets rank 2, etc. Next, the rank of each (shortest) derivation is computed as the sum of the ranks of the subtrees involved. The derivation with the smallest sum, or highest rank, is taken as the best derivation producing the best parse tree.

The way we compute the rank of a derivation by summing up the ranks of its subtrees may seem rather ad hoc. However, it is possible to provide an information-theoretical motivation for this model. According to Zipf's law, rank is roughly proportional to the negative logarithm of frequency (Zipf 1935). In Shannon's Information Theory (Shannon 1948), the negative logarithm (of base 2) of the probability of an event is better known as the *information* of that event. Thus, the rank of a subtree is roughly proportional to its information. It follows that minimizing the sum of the subtree ranks in a derivation corresponds to minimizing the (self-)information of a derivation.

## 3. Computational Aspects

### 3.1 Computing the most probable parse

Bod (1993) showed how standard chart parsing techniques can be applied to probabilistic DOP. Each corpus-subtree $t$ is converted into a context-free rule $r$ where the lefthand side of $r$ corresponds to the root label of $t$ and the righthand side of $r$ corresponds to the frontier labels of $t$. Indices link the rules to the original subtrees so as to maintain the subtree's internal structure and probability. These rules are used to create a derivation forest for a sentence, and the most probable parse is computed by sampling a sufficiently large number of random derivations from the forest ("Monte Carlo disambiguation", see Bod 1998; Chappelier & Rajman 2000). While this technique has been successfully applied to parsing the ATIS portion in the Penn Treebank (Marcus et al. 1993), it is extremely time consuming. This is mainly because the number of random derivations that should be sampled to reliably estimate the most probable parse increases exponentially with the sentence length (see Goodman 1998). It is therefore questionable whether Bod's sampling technique can be scaled to larger corpora such as the OVIS and the WSJ corpora.

Goodman (1998) showed how the probabilistic DOP model can be reduced to a compact stochastic context-free grammar (SCFG) which contains exactly eight SCFG rules for each node in the training set trees. Although Goodman's reduction method does still not allow for an efficient computation of the most probable parse in DOP (in fact, the problem of computing the most probable parse is NP-hard -- see Sima'an 1996), his method does allow for an efficient computation of the "maximum constituents parse", i.e., the parse tree that is most likely to have the largest number of correct constituents (also called the "labeled recall parse"). Goodman has shown on the ATIS corpus that the maximum constituents parse performs at least as well as the most probable parse if all subtrees are used. Unfortunately, Goodman's reduction method remains beneficial only if indeed *all* treebank subtrees are used (see Sima'an 1999: 108), while maximum parse accuracy is typically obtained with a subtree set which is smaller than the total set of subtrees (this is probably due to data-sparseness effects -- see Bonnema et al. 1997; Bod 1998; Sima'an 1999).

In this paper we will use Bod's subtree-to-rule conversion method for studying the behavior of probabilistic against non-probabilistic DOP for different maximum subtree sizes. However, we will not use Bod's Monte Carlo sampling technique from complete derivation forests, as this turns out to be computationally impractical for our larger corpora. Instead, we use a Viterbi *n*-best search and estimate the most probable parse from the 1,000 most probable derivations, summing up the probabilities of derivations that generate the same tree. The algorithm for computing *n* most probable derivations follows straightforwardly from the algorithm which computes the most probable derivation by means of Viterbi optimization (see Sima'an 1995, 1999).

### 3.2 Computing the shortest derivation

As with the probabilistic DOP model, we first convert the corpus-subtrees into rewrite rules. Next, the shortest derivation can be computed in the same way as the most probable derivation (by Viterbi) if we give all rules equal probabilities, in which case the shortest derivation is equal to the most probable derivation. This can be seen as follows: if each rule has a probability $p$ then the probability of a derivation involving $n$ rules is equal to $p^n$, and since $0<p<1$ the derivation with the fewest rules has the greatest probability. In our experiments, we gave each rule a probability mass equal to $1/R$, where $R$ is the number of distinct rules derived by Bod's method.

As mentioned above, the shortest derivation may not be unique. In that case we compute *all* shortest derivations of a sentence and then apply our ranking scheme to these derivations. Note that this ranking scheme *does* distinguish between subtrees or different root labels, as it ranks the subtrees given their root label. The ranks of the shortest derivations are computed by summing up the ranks of the subtrees they involve. The shortest derivation with the smallest sum of subtree ranks is taken to produce the best parse tree.[3]

---

[3] It may happen that different shortest derivations generate the same tree. We will not distinguish between these cases, however, and compute only the shortest derivation with the highest rank.

## 4. Experimental Comparison

### 4.1 Experiments on the ATIS corpus

For our first comparison, we used 10 splits from the Penn ATIS corpus (Marcus et al. 1993) into training sets of 675 sentences and test sets of 75 sentences. These splits were random except for one constraint: that all words in the test set actually occurred in the training set. As in Bod (1998), we eliminated all epsilon productions and all "pseudo-attachments". As accuracy metric we used the exact match defined as the percentage of the best parse trees that are identical to the test set parses. Since the Penn ATIS portion is relatively small, we were able to compute the most probable parse both by means of Monte Carlo sampling and by means of Viterbi *n*-best. Table 1 shows the means of the exact match accuracies for increasing maximum subtree depths (up to depth 6).

| Depth of subtrees | Probabilistic DOP | | Non-probabilistic DOP |
|---|---|---|---|
| | Monte Carlo | Viterbi *n*-best | |
| 1 | 46.7 | 46.7 | 24.8 |
| ≤2 | 67.5 | 67.5 | 40.3 |
| ≤3 | 78.1 | 78.2 | 57.1 |
| ≤4 | 83.6 | 83.0 | 81.5 |
| ≤5 | 83.9 | 83.4 | 83.6 |
| ≤6 | 84.1 | 84.0 | 85.6 |

Table 1. Exact match accuracies for the ATIS corpus

The table shows that the two methods for probabilistic DOP score roughly the same: at depth ≤ 6, the Monte Carlo method obtains 84.1% while the Viterbi *n*-best method obtains 84.0%. These differences are not statistically significant. The table also shows that for small subtree depths the non-probabilistic DOP model performs considerably worse than the probabilistic model. This may not be surprising since for small subtrees the shortest derivation corresponds to the smallest parse tree which is known to be a bad prediction of the correct parse tree. Only if the subtrees are larger than depth 4, the non-probabilistic DOP model scores roughly the same as its probabilistic counterpart. At subtree depth ≤ 6, the non-probabilistic DOP model scores 1.5% better than the best score of the probabilistic DOP model, which is statistically significant according to paired *t*-tests.

### 4.2 Experiments on the OVIS corpus

For our comparison on the OVIS corpus (Bonnema et al. 1997; Bod 1998) we again used 10 random splits under the condition that all words in the test set occurred in the training set (9000 sentences for training, 1000 sentences for testing). The OVIS trees contain both syntactic and semantic annotations, but no epsilon productions. As in Bod (1998), we treated the syntactic and semantic annotations of each node as one label. Consequently, the labels are very restrictive and collecting statistics over them is difficult. Bonnema et al. (1997) and Sima'an (1999) report that (probabilistic) DOP suffers considerably from data-sparseness on OVIS, yielding a decrease in parse accuracy if subtrees larger than depth 4 are included. Thus it is interesting to investigate how non-probabilistic DOP behaves on this corpus. Table 2 shows the means of the exact match accuracies for increasing subtree depths.

| Depth of subtrees | Probabilistic DOP | Non-probabilistic DOP |
|---|---|---|
| 1 | 83.1 | 70.4 |
| ≤2 | 87.6 | 85.1 |
| ≤3 | 89.6 | 89.5 |
| ≤4 | 90.0 | 90.9 |
| ≤5 | 89.7 | 91.5 |
| ≤6 | 88.8 | 92.2 |

Table 2. Exact match accuracies for the OVIS corpus

We again see that the non-probabilistic DOP model performs badly for small subtree depths while it outperforms the probabilistic DOP model if the subtrees get larger (in this case for depth > 3). But while the accuracy of probabilistic DOP deteriorates after depth 4, the accuracy of non-probabilistic DOP continues to grow. Thus non-probabilistic DOP seems relatively insensitive to the low frequency of larger subtrees. This property may be especially useful if no meaningful statistics can be collected while sentences can still be parsed by large chunks. At depth ≤ 6, non-probabilistic DOP scores 3.4% better than probabilistic DOP, which is statistically significant using paired $t$-tests.

### 4.3 Experiments on the WSJ corpus

Both the ATIS and OVIS corpus represent restricted domains. In order to extend our results to a broad-coverage domain, we tested the two models also on the Wall Street Journal portion in the Penn Treebank (Marcus et al. 1993).

To make our results comparable to others, we did not test on different random splits but used the now standard division of the WSJ with sections 2-21 for training (approx. 40,000 sentences) and section 23 for testing (see Collins 1997, 1999; Charniak 1997, 2000; Ratnaparkhi 1999); we only tested on sentences ≤ 40 words (2245 sentences). All trees were stripped off their semantic tags, co-reference information and quotation marks. We used all training set subtrees of depth 1, but due to memory limitations we used a subset of the subtrees larger than depth 1 by taking for each depth a random sample of 400,000 subtrees. No subtrees larger than depth 14 were used. This resulted in a set of 5,217,529 subtrees which were smoothed by the technique described in Bod (1996). We did not employ a separate part-of-speech tagger: the test sentences were directly parsed by the training set subtrees. For words that were unknown in the training set, we guessed their categories by means of the method described in Weischedel et al. (1993) which uses statistics on word-endings, hyphenation and capitalization. The guessed category for each unknown word was converted into a depth-1 subtree and assigned a probability (or frequency for non-probabilistic DOP) by means of simple Good-Turing.

As accuracy metric we used the standard PARSEVAL scores (Black et al. 1991) to compare a proposed parse $P$ with the corresponding correct treebank parse $T$ as follows:

$$\text{Labeled Precision} = \frac{\text{\# correct constituents in } P}{\text{\# constituents in } P}$$

$$\text{Labeled Recall} = \frac{\text{\# correct constituents in } P}{\text{\# constituents in } T}$$

A constituent in $P$ is "correct" if there exists a constituent in $T$ of the same label that spans the same words. As in other work, we collapsed ADVP and PRT to the same label when calculating these scores (see Collins 1997; Ratnaparkhi 1999; Charniak 1997).

Table 3 shows the labeled precision (LP) and labeled recall (LR) scores for probabilistic and non-probabilistic DOP for six different maximum subtree depths.

| Depth of subtrees | Probabilistic DOP | | Non-probabilistic DOP | |
|---|---|---|---|---|
| | LP | LR | LP | LR |
| ≤4 | 84.7 | 84.1 | 81.6 | 80.1 |
| ≤6 | 86.2 | 86.0 | 85.0 | 84.7 |
| ≤8 | 87.9 | 87.1 | 87.2 | 87.0 |
| ≤10 | 88.6 | 88.0 | 86.8 | 86.5 |
| ≤12 | 89.1 | 88.8 | 87.1 | 86.9 |
| ≤14 | 89.5 | 89.3 | 87.2 | 86.9 |

Table 3. Scores on the WSJ corpus (sentences ≤ 40 words)

The table shows that probabilistic DOP outperforms non-probabilistic DOP for maximum subtree depths 4 and 6, while the models yield rather similar results for maximum subtree depth 8. Surprisingly, the scores of non-probabilistic DOP deteriorate if the subtrees are further enlarged, while the scores of probabilistic DOP continue to grow, up to 89.5% LP and 89.3% LR. These scores are higher than those of several other parsers (e.g. Collins 1997, 99; Charniak 1997), but remain behind the scores of Charniak (2000) who obtains 90.1% LP and 90.1% LR for sentences ≤ 40 words. However, in Bod (2000b) we show that even higher scores can be obtained with probabilistic DOP by restricting the number of words in the subtree frontiers to 12 and restricting the depth of unlexicalized subtrees to 6; with these restrictions an LP of 90.8% and an LR of 90.6% is achieved.

We may raise the question as to whether we actually need these extremely large subtrees to obtain our best results. One could argue that DOP's gain in parse accuracy with increasing subtree depth is due to the model becoming sensitive to the influence of lexical heads higher in the tree, and that this gain could also be achieved by a more compact depth-1 DOP model (i.e. an SCFG) which annotates the nonterminals with headwords. However, such a head-lexicalized stochastic grammar does not capture dependencies between nonheadwords (such as *more* and *than* in the WSJ construction *carry more people than cargo* where neither *more* nor *than* are headwords of the NP-constituent *more people than cargo*), whereas a frontier-lexicalized DOP model using large subtrees does capture these dependencies since it includes subtrees in which e.g. *more* and *than* are the only frontier words. In order to isolate the contribution of nonheadword dependencies, we eliminated all subtrees containing two or more nonheadwords (where a nonheadword of a subtree is a word which is not a headword of the subtree's root nonterminal -- although such a nonheadword may be a headword of one of the subtree's internal nodes). On the WSJ this led to a decrease in LP/LR of 1.2%/1.0% for probabilistic DOP. Thus nonheadword dependencies contribute to higher parse accuracy, and should not be discarded. This goes against common wisdom that the relevant lexical dependencies can be restricted to the locality of headwords of constituents (as advocated in Collins 1999). It also shows that DOP's frontier lexicalization is a viable alternative to constituent lexicalization (as proposed in Charniak 1997; Collins 1997, 99; Eisner 1997). Moreover, DOP's use of large subtrees makes the model not only more lexically but also more structurally sensitive.

## 5. Conclusion

Common wisdom has it that the bias of stochastic grammars in favor of shorter derivations is harmful and should be redressed. We have shown that the common wisdom is wrong for stochastic tree-substitution grammars that use elementary trees of flexible size. For such grammars, a non-probabilistic metric based on the shortest derivation outperforms a probabilistic metric on the ATIS and OVIS corpora, while it obtains competitive results on the Wall Street Journal corpus. We have seen that a non-probabilistic version of DOP performed especially well on corpora for which collecting subtree statistics is difficult, while sentences can still be parsed by relatively large chunks. We have also seen that probabilistic DOP obtains very competitive results on the WSJ corpus. Finally, we conjecture that any stochastic grammar which uses elementary trees rather than context-free rules can be turned into an accurate non-probabilistic version (e.g. Tree-Insertion Grammar and Tree-Adjoining Grammar).

## Acknowledgements

Thanks to Khalil Sima'an and three anonymous reviewers for useful suggestions.